\documentclass[letterpaper]{article} %DO NOT CHANGE THIS
\usepackage{aaai19}  %Probably error in formatting instructions. Style File is called aaai19.sty -> renamed to aaai.sty
\usepackage{times}  %Required
\usepackage{helvet}  %Required
\usepackage{courier}  %Required
\usepackage{url}  %Required
\usepackage{graphicx}  %Required
 \usepackage{amssymb}
 \usepackage{booktabs} % nicer Tables
 \usepackage{multirow} % nicer Tables

\title{An Open-World Extension to Knowledge Graph Completion Models}
\author{Haseeb Shah,\textsuperscript{1}
Johannes Villmow,\textsuperscript{2}
Adrian Ulges,\textsuperscript{2}
Ulrich Schwanecke,\textsuperscript{2}
Faisal Shafait\textsuperscript{1}\\
\textsuperscript{1}{National University of Science and Technology, Pakistan}\\
\textsuperscript{2}{RheinMain University of Applied Sciences, Germany}\\
\{hshah.bese15seecs, faisal.shafait\}@seecs.edu.pk\\
\{johannes.villmow, adrian.ulges, ulrich.schwanecke\}@hs-rm.de}

\begin{document}
\maketitle

\begin{abstract}
We present a novel extension to embedding-based knowledge graph completion models which enables them to perform open-world link prediction, i.e. to predict facts for entities unseen in training based on their textual description. 
Our model combines a regular link prediction model learned from a knowledge graph with word embeddings learned from a textual corpus. After training both independently, we learn a transformation to map the embeddings of an entity's name and description to the graph-based embedding space. 

In experiments on several datasets including FB20k, DBPedia50k and our new dataset FB15k-237-OWE, we demonstrate competitive results.
Particularly, our approach exploits the full knowledge graph structure even when textual descriptions are scarce, does not require a joint training on graph and text, and can be applied to any embedding-based link prediction model, such as TransE, ComplEx and DistMult. 
%We study different transformations, word embeddings and link prediction models, and 
\end{abstract}

%even in knowledge graphs with scarce textual descriptions,  -- our approach makes use of the full graph structure and achieves robust learning. 

\section{Introduction}
Knowledge graphs are a vital source for disambiguation and discovery in various tasks such as question answering~\cite{ferrucci10watson}, information extraction~\cite{dong14knowledgevault} and search~\cite{gogle12knowledgegraph}. They are, however, known to suffer from data quality issues~\cite{paulheim17kgrefinement}. Most prominently, since formal knowledge is inherently sparse, relevant facts are often missing from the graph. 

To overcome this problem, {\it knowledge graph completion} (KGC) or \textit{link prediction} strives to enrich existing graphs with new facts. Formally, a knowledge graph  $\mathcal{G} \, {\subset} \, E {\times} R {\times} E$ consists of facts or triples $(head,rel,tail)$, where $E$ and $R$ denote finite sets of entities and relations respectively. Knowledge graph completion is targeted at assessing the probability of triples not present in the graph.
To do so, a common approach involves representing the entities and relations in triples using real-valued vectors called embeddings. The probability of the triple is then inferred by geometric reasoning over the embeddings. Embeddings are usually generated by learning to discriminate real triples from randomly corrupted ones~\cite{nickel16kg,shi16proje,trouillon2016complex}. 

A key problem with most existing approaches is that the plausibility of links can be determined for known entities only. For many applications, however, it is of interest to infer knowledge about entities not present in the graph. Imagine answering the question ``What is German actress \textit{Julia Lindig} known for?'', where \textit{Julia Lindig} is not a known entity. Here, information can be inferred from the question, typically using word embeddings~\cite{mikolov13w2v,Pennington14glove:global,peters18elmo}. Similar to entity embeddings, these techniques represent words with embedding vectors. These can be pre-trained on text corpora, thereby capturing word similarity and semantic relations, which may help to predict the plausibility of the triple $(Julia\_Lindig, starred\_in, Lola\_Rennt)$. This challenge is known as open-world (or zero-shot) KGC. To the best of our knowledge, few open-world KGC models have been proposed so far, all of which are full replacements for regular KGC models and require textual descriptions for all entities~\cite{xie16descriptions,shi17openworld}.

In this paper, we suggest a different approach, namely to extend existing KGC models with pre-trained word embeddings.  Given an new entity, we aggregate its name and description into a text-based entity representation. We then learn a transformation from text-based embedding space to graph-based embedding space, where we can now apply the graph-based model for predicting links.
We show that this simple approach yields competitive results, and offers two key benefits: First, it is independent of the specific KGC model used, which allows us to use multiple different link prediction models from which we can pick the best one. Second, as training on the graph structure happens independently from training on text, our approach can exploit the full-scale knowledge graph structure in situations where textual information is scarce because learning the transformation is robust even for such situations. We coin our approach OWE for Open World Extension and combine it with several common KGC models, obtaining TransE-OWE, DistMult-OWE, and ComplEx-OWE.

We demonstrate competitive results on common datasets for open-world prediction, and also introduce a new dataset called FB15k-237-OWE, which avoids bias towards long textual descriptions and trivial regularities like inverse relations. 
% Beyond this, we study the building blocks of our model (KGC model, text embeddings, and the transformations) in depth. 
The code and the new FB15k-237-OWE dataset are available online\footnote{\url{https://github.com/haseebs/OWE}}.

%Whichever approach is taken for constructing a
%knowledge graph, the result will never be perfect [10].
%As a model of the real world or a part thereof, formal-
%ized knowledge cannot reasonably reach full coverage,
%i.e., contain information about each and every entity in
%the universe. Furthermore, it is unlikely, in particular
%when heuristic methods are applied, that the knowl-
%edge graph is fully correct – there is usually a trade-off
%between coverage and correctness, which is a

\section{Related Work} \label{relatedwork}
\begin{figure*}[t!]
\centering
\includegraphics[width=\textwidth]{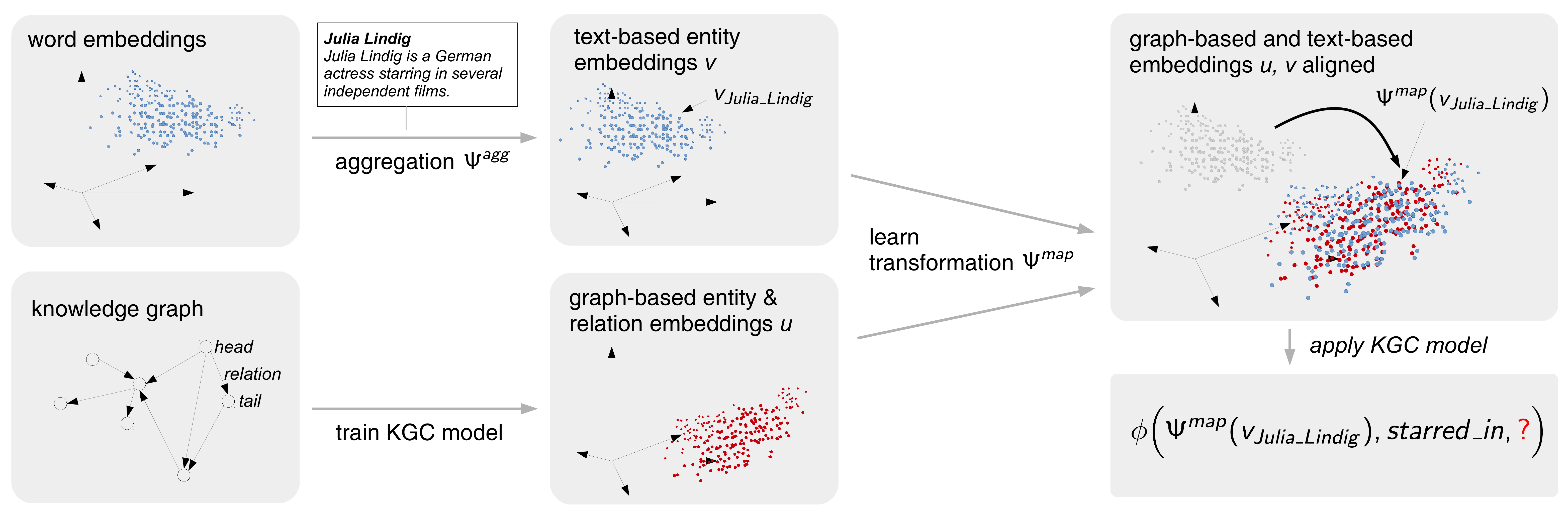}
\caption{Our approach first trains a KGC model on the graph without using textual information (bottom left). For every entity we can obtain a text-based embedding $v$ by aggregating the word embeddings for tokens in the name and description (top left). A transformation $\Psi^{map}$ is learned on the training entities to map $v$ to the space of graph-based embeddings (right). The learned mapping can then be applied to unknown entities, thus allowing the trained KGC model to be applied.}
\label{fig:approach}
\end{figure*} 
%%Figure 1: Open-world link prediction: Word embeddings are used to derive a text-based embeddingvfor a new entity unknownto the knowledge graph (top left). A transformationΨmapis learned to mapvto the space of graph-based embeddings (right),where the trained knowledge graph completion (KGC) model can be applied

% Some related papers: https://github.com/thunlp/KRLPapers
\subsubsection{Knowledge Graph Completion} Interest in KGC has increased in the past few years, with a focus on embedding-based methods. A concise survey of earlier works such as NTN \cite{socher13ntn} and TransE \cite{bordes2013transe} is provided by \cite{nickel16kg}. TransE has been recently complemented by other models like DistMult \cite{yang2015distmult}, ComplEx \cite{trouillon2016complex}, ProjE~\cite{shi16proje} and RDF2Vec~\cite{ristoski16rdf2vec}.

A common approach is to estimate the probability of triples $(head,rel,tail)$ using a scoring function
$
 \phi(u_{head}, u_{rel}, u_{tail}),
$
where $u_{x}$ denotes the embedding of entity/relation $x$  and is a real-valued or complex-valued vector. $\phi$ depends on the model and varies from simple translation~\cite{bordes2013transe} over bilinear forms~\cite{yang2015distmult} to complex-valued forms~\cite{trouillon2016complex}. 
Training happens by randomly perturbing triples in the graph 
and learning to discriminate real triples from perturbed ones, typically by using negative sampling.

%Maybe Poincaré Embeddings for Learning Hierarchical Representations ?

\subsubsection{Word Embeddings} Embedding-based representations of text have become a common approach in natural language processing~\cite{goldberg16primer}. They represent terms, sentences or documents in form of a vector.  
Word embeddings are known to capture term similarities and semantic relations~\cite{mikolov13w2v,Pennington14glove:global}. Word embeddings can also include sub-word information for out-of-training-vocabulary generalization~\cite{bojanowski16enriching,peters18elmo}, and have been combined with anchor and link information obtained from Wikipedia to produce entity-specific embeddings~\cite{wikipedia2vec}.

\subsubsection{Text-Enhanced Knowledge Graph Completion}
While the knowledge graph completion models described above leverage only the triple structure of the graph, some approaches combine text information with the graph information. 
Some of these approaches regularize word embeddings based on semantic information, for example, by adding synonym and other relations from WordNet to the training set of contextual term pairs~\cite{faruqui14retrofitting,yu14improving} or by modelling relations like synonyms with translations~\cite{xu14rcnet}. These methods, however, are not targeted at KGC.

Closer to our work are actual KGC models that employ text information, such that for entities scarcely linked in the graph, extra information can be drawn from text.
\cite{socher13ntn,wang16teke} use averaged pre-trained word vectors (CBOW) as entity descriptions for geometric reasoning with affine mappings. \cite{xu17jointly} tests different aggregation functions for word embeddings (CBOW, LSTMs, attentive LSTMs) and proposes learning an entity-wise linear interpolation between graph-based and text-based embedding, which is combined with a translational KGC model.
\cite{toutanova2015fb15k237} enhance KGC with a large set of relations extracted from dependency parses on a text corpus, and formulate a joint loss function for text-based and graph-based inputs. However, none of these works address the open-world setting that we target in this paper.

Only few other works address open-world KGC. {\it Description-Embodied Knowledge Representation Learning (DKRL)}~\cite{xie16descriptions} uses a joint training of graph-based embeddings (TransE) and text-based embeddings while regularizing both types of embeddings to be aligned using an additional loss. 
ConMask \cite{shi17openworld} is a text-centric approach where text-based embeddings for head, relation and tail are derived by an attention model over names and descriptions, and the triple $(h,r,t)$ is scored by a pairwise matching of the embeddings, followed by a softmax regression.
In contrast to these approaches, we train graph and text embeddings independently. This comes with two benefits: First, our approach can fully leverage knowledge graphs with incomplete or scarce textual descriptions. Second, it is applicable to any embedding-based link prediction model such as TransE, DistMult and ComplEx.
%Second, a selective refinement of those terms in the training graph -- which may give rise to overfitting -- is avoided.  
% Second, refining pre-trained word embeddings leads to a discrepancy between terms that are found in the training graph and similar terms that are not. Our approach of mapping from text-based space to graph-based space avoids such overfitting.

%{\bf Knowledge Graphs in Information Extraction}
%... only 
%The combination of text sources and knowledge graphs %has also been studied in information extraction. 

\section{Approach} \label{approach}
Our approach starts with a regular link prediction model (in the following also referred to as the {\it graph-based} model) as outlined in Section \ref{relatedwork} and visualised in Fig. \ref{fig:approach}. The model scores triples $(h, r, t)$:

\begin{equation}
\label{eq:kg_scoring}
score(h,r,t) = \phi({u}_{h}, {u}_{r}, {u}_{t}) 
\end{equation}

%where $\mathbf{u}_{x}$ denotes the embedding of entity/relation $x$. Typically,  $\mathbf{u}_{x} \in \mathbb{R}^d$ is a real-valued vector of a few hundred dimensions, but other options are possible (e.g., in ComplEx~\cite{trouillon2016complex}, $\mathbf{u}_{x} \in \mathbb{C}^d$). $\phi^{graph}$ is a scoring function that depends on the model and varies from simple shifts over scalar products to affine or bilinear models.

where ${u}_{x}$ denotes the embedding of entity/relation $x$. Typically,  ${u}_{x} \in \mathbb{R}^d$, but other options are possible. For example, in ComplEx~\cite{trouillon2016complex}, ${u}_{x}$ is complex-valued (${u}_{x} \in \mathbb{C}^d$). $\phi$ is a scoring function that depends on the link prediction model and will be adressed in more detail in Section \ref{kgcmodels}. %varies from linear shifts over scalar products to affine or bilinear models.
\subsubsection{Closed-world Link Prediction}Closed-world link prediction involves predicting facts about entities on which the link prediction model is trained on. For tail prediction, head and relation are given and the objective is to predict the tail entity with the highest score. This is done by calculating the score of the $(h,r)$ pair with every $t \in E$.

\begin{equation}
tail^* = \arg\max_{t \in E} \;\; score(h, r, t)
\end{equation}
Similarly, for head prediction, the score of $(r,t)$ is calculated with every $h \in E$:
\begin{equation}
head^* = \arg\max_{h \in E} \;\; score(h, r, t)
\end{equation}
For the remaining part of the paper, we will only discuss the task of tail prediction. The same concepts can be applied to the task of head prediction.
%, though he same concepts can be generalized to head prediction.

%This is done by employing the above graph-based model: For example, given a known entity $head \in E$ and relation $rel \in R$, the above model estimates the best-matching tail  $(head,rel,?)$ using Equation \ref{eq:kg_scoring}:
%$$
%tail^* = \arg\max_{tail \in E} (\phi^{graph}(\mathbf{u}_{head}, \mathbf{u}_{rel}, \mathbf{u}_{tail}))
%$$
%Note that predicting tails $(?,rel,tail)$ works the same way.

\subsubsection{Open-world Extension} Open-world link prediction involves predicting facts about  entities that the link prediction model was not trained on. Our contribution lies in extending the above graph-based model to perform open-world link prediction. We assume an unseen head entity $head \not\in E$ to be represented by its name and textual description, which we concatenated into a word sequence $\mathcal W = (w_1, w_2, ...,w_n)$. 
Word embeddings (such as Word2Vec or Glove) pre-trained on a text corpus are then used to transform the sequence of tokens $\mathcal W$ into a sequence of embeddings $({v}_{w_1}, {v}_{w_2},...,{v}_{w_n})$. This sequence of embeddings is then aggregated with a function $\Psi^{agg}$ to obtain a text-based embedding of the head entity ${v}_{h} \in \mathbb{R}^{d'}$:
\begin{equation}
\label{eq:aggregation}
{v}_{h} := \Psi^{agg} \Big( {v}_{w_1}, {v}_{w_2}, ..., {v}_{w_n} \Big) 
\end{equation}
Since text-based and graph-based embeddings are trained independently on different information sources, we cannot expect them to match. Therefore, a transformation $\Psi^{map}$ is learned   from text-based embedding space to graph-based embedding space such that $\Psi^{map}({v}_{h}) \approx {u}_{h}$. Then we score triples with the unseen head entity by applying the graph-based model from Equation \ref{eq:kg_scoring} with the mapped text-based head description:
\begin{equation}
score(h, r, t) = \phi\Big(\Psi^{map}({v}_{h}), {u}_{r}, {u}_{t}\Big)
\end{equation}
The single steps of our model are outlined in more detail in the following.

\subsection{Link Prediction Models} \label{kgcmodels}
Since our approach is independent of the specific link prediction model used, we test three commonly used models in this work:
%(1) TransE~\cite{bordes2013transe}
\begin{enumerate}
    \item TransE: $\phi({u}_{h}, {u}_{r}, {u}_{t}) = -||{u}_{h} {+} {u}_{r} {-} {u}_{t}||_2$ $\;\;\;$ 
    \item DistMult: $\phi({u}_{h}, {u}_{r}, {u}_{t}) = \langle{u}_{h},{u}_{r},{u}_{t}\rangle$ $\;\;\;\;\;\;\;\;$ 
    \item ComplEx: $\phi({u}_{h}, {u}_{r}, {u}_{t}) = \textrm{Re}({\langle}{u}_{h}, {u}_{r}, {\overline{u}}_{t}{\rangle})$ $\;$ 
\end{enumerate}
Note that the first two use real-valued embeddings, while ComplEx uses complex-valued embeddings (where ${\overline{u}} = \textrm{Re}({u}) - i{\cdot} \textrm{Im}({u})$ denotes the complex conjugate of embedding ${u}$).
All models are trained using their original loss functions and validated using closed-world validation data.
%Free parameters (the amount of negative samples and learning rate) were optimized on hold-out validation data.

\subsection{Word Embeddings and Aggregation}
\label{sec:word_embeddings}

We use pre-trained word embeddings trained on large text corpora. Since the number of entities in the datasets used is limited and we found overfitting to be an issue, we omit any refinement of the embeddings. We tested  200-dimensional Glove embeddings~\cite{Pennington14glove:global} and 300-dimensional Wikipedia2Vec  embeddings~\cite{wikipedia2vec}. 

Note that Wikipedia2Vec embeddings contain phrase embeddings, which we use as an embedding for entity names (like "Julia Lindig"). If no phrase embedding is available, we split the name into single tokens and use token-wise embeddings. If no embedding is available for a token, we use a vector of zeros as an ``unknown'' token.

To aggregate word embeddings to an entity embedding (function $\Psi^{agg}$, Equation \ref{eq:aggregation}), approaches in the literature range from simple averaging~\cite{Pennington14glove:global} over Long Short Term Memory Networks (LSTMs)~\cite{xu17jointly} to relation-specific masking~\cite{shi17openworld}. We use averaging as an aggregation function. Here, the word embedding vectors are averaged to obtain a single representative embedding. 
To prevent overfitting, we apply dropout during training, i.e. embeddings of some words are randomly replaced by the unknown token before averaging.

\subsection{Transformation Functions}
The key to open-world prediction is the mapping from text-based entity embeddings ${v}_e$ to graph-based ones ${u}_e$. Several different transformation functions $\Psi^{map}$ can be learned for this task. In this paper, we discuss three options:

\subsubsection{Linear} A simple linear function $\Psi^{map}({v}) = A {\cdot} {v}$. For ComplEx, separate matrices are used for the real and imaginary part: $\Psi^{map}({v}) = A {\cdot} {v} + i \cdot A' {\cdot} {v}$

\subsubsection{Affine} Here, $\Psi^{map}$ is an affine function $\Psi^{map}({v}) = A {\cdot} {v} + {b}$. For ComplEx, separate matrices and vectors are trained just like above: $\Psi^{map}({v}) = (A {\cdot} {v} + {b}) + i \cdot (A' {\cdot} {v} + {b'})$

\subsubsection{MLP} $\Psi^{map}$ is a four layer Multi-Layer Perceptron (MLP) with ReLU activation functions. The output layer is affine. We did not perform an extensive hyperparameter search here.

To train the transformations, first a link prediction model is trained on the full graph, obtaining entity embeddings ${u}_1,...,{u}_n$. 
We then choose all entities $e_{i_1},...,e_{i_m}$ with textual metadata (names and/or descriptions), and extract text-based embedding ${v}_{i_1},...,{v}_{i_m}$ for them using aggregation (see above). Finally, $\Psi^{map}$ is learned by minimizing the loss function 
\begin{equation}
      L(\Theta) = \sum_{k=1}^m \Big| \Big|\Psi_\Theta^{map}({v}_{i_k}) - {u}_{i_k} \Big|\Big|_2
%  E(\Theta) = \sum_{i=1}^n 1 - cos()
\end{equation}
using batched stochastic gradient descent, where $\Theta$ denotes the parameters of $\Psi^{map}$ (e.g., the weight matrices and bias vectors $A,b$). For ComplEx, the above loss is summed for real and imaginary parts, and training happens on the sum. We apply no fine-tuning, neither on the graph nor on the text embeddings.

\section{Experiments} \label{experiments}
In this section, we study the impact of our model's parameters ($\Psi^{agg}, \Psi^{map}$, text embeddings) on prediction performance. We also provide mappings of selected open-world entities, and compare our results with the state-of-the-art.

\begin{table}[h]
\centering
\begin{tabular}{@{\extracolsep{-1.3ex}}lrrrrr@{}}
	\toprule \\[-0.8em]
\multicolumn{1}{l}{} & \multicolumn{1}{c}{} & \multicolumn{1}{c}{} & \multicolumn{3}{c}{{Number of Triples}} \\ \cmidrule(l){4-6} 
\multicolumn{1}{c}{{Dataset}} & \multicolumn{1}{c}{{$|E|$}} & \multicolumn{1}{c}{{$|R|$}} & \multicolumn{1}{c}{{Train}} & \multicolumn{1}{c}{{Valid}} & \multicolumn{1}{c}{{Test}} \\ \midrule
FB15k / FB20k & 14,904 & 1,341 & 472,860 & 48,991 & 57,803 \\
% FB20k & 19,923 & 1,341 & 472,860 & 48,991 & 90,149 \\
DBPedia50k & 24,624 & 351 & 32,388 & 123 & 2,095 \\
FB15k-237-OWE & 12,324 & 235 & 242,489 & 12,806 & - \\ 	\bottomrule
\end{tabular}
\caption{Dataset statistics for closed-world link prediction.}
\label{table:closed-world-datasets}
\end{table}
\begin{table}[h]
\begin{tabular}{@{\extracolsep{-1ex}}lrrrrr@{}}
	\toprule \\[-0.8em]
\multicolumn{1}{l}{} & \multicolumn{1}{c}{} & \multicolumn{2}{c}{{Head Pred.}} & \multicolumn{2}{l}{{Tail Pred.}} \\ \cmidrule(l){3-4} \cmidrule(l){5-6} 
\multicolumn{1}{c}{{Dataset}} & \multicolumn{1}{c}{$|E^{open}|$} & \multicolumn{1}{c}{{Valid}} & \multicolumn{1}{c}{{Test}} & {Valid} & {Test} \\ \midrule
FB20k & 5,019 & - & 18,753 & - & 11,586 \\
DBPedia50k & 3,636 & 55 & 2,139 & 164 & 4,320 \\
FB15k-237-OWE & 2,081 & 1,539 & 13,857 & 9,424 & 22,393 \\\bottomrule 
\end{tabular}
\caption{Dataset statistics for open-world link prediction. $E^{open}$ is the set of novel entities not in the graph. Note that the corresponding closed-world training set is used for training the open-world models.}
\label{table:open-world-datasets}
\end{table}

\subsection{Datasets}

Closed-world KGC tasks are commonly evaluated on WordNet and Freebase subsets, such as WN18, WN18RR, FB15k, and FB15k-237.
For open-world KGC, the following datasets have been suggested: \cite{xie16descriptions} introduced FB20k, which builds upon the FB15k dataset by  adding test triples with unseen entities, which are selected to have long textual descriptions.
\cite{shi17openworld} introduced DBPedia50k and DBPedia500k datasets for both open-world and closed-world KGC tasks. 

However, the above datasets display a bias towards long textual descriptions: DBpedia50k has an average description length of 454 words, FB20k of 147 words.
Also, for neither of the datasets precautions have been taken to avoid redundant inverse relations, which allows models to exploit trivial patterns in the data~\cite{toutanova2015fb15k237}. To overcome these problems, we introduce a new dataset named FB15k-237-OWE. FB15k-237-OWE is based on the well-known FB15K-237 dataset, where redundant inverse relations have been removed. Also, we avoid a bias towards entities with longer textual descriptions: Test entities are uniformly sampled from FB15K-237, and only short Wikidata descriptions (5 words on average) are used. 

In the following section, the sampling strategy for FB15k-237-OWE is briefly outlined: For tail prediction test set, we start with FB15K-237 and randomly pick heads (by uniform sampling over all head entities). Each picked head $x$ is removed from the training graph by moving all triples of the form $(x,?,t)$ to the test set and dropping all triples of the form $(?,?,x)$ if $t$ still remains in the training set after these operations. Similarly, a head prediction test set is prepared from the set of dropped triplets which satisfy the conditions to be in head prediction test set i.e. head must be represented in training set while tail must not be represented.
The dataset also contains two validation sets: A closed-world one (with random triples picked from the training set) and an open-world one (with random triples picked from the test set).

%We introduce a new dataset FB15k-237-OWE (FB237ZS) for open-world link prediction, which is generated using FB15k-237. 
We evaluate our approach on DBPedia50k, FB20k, and the new dataset FB15k-237-OWE. Statistics of the datasets are highlighted in Table \ref{table:closed-world-datasets} and Table \ref{table:open-world-datasets}. 
%Note that for DBPedia50k, we found the number of entities/relations used by the published evaluation code to be lower than claimed in~\cite{shi17openworld}.

\begin{table*}[t]
	\centering
	\begin{tabular}{@{\extracolsep{0.15ex}}l r r r r r r r r r r r r} 
		\toprule \\[-0.8em]
		& \multicolumn{4}{c}{DBPedia50k} & \multicolumn{4}{c}{FB15k-237-OWE} & \multicolumn{4}{c}{FB20k} \\
		\cline{2-5} \cline{6-9} \cline{10-13} \\[-0.8em] 
		\multicolumn{1}{c}{Model} & H@1 & H@3 & H@10 & MRR & H@1 & H@3 & H@10 & MRR & H@1 & H@3 & H@10 & MRR\\
		\midrule \\[-0.8em]
		Target Filt. Base. & 4.5 & 9.7 & 23.0 & 11.0* & 6.4 & 14.2 & 23.3 & 12.7 & 17.5 & 32.1 & 41.2 & 27.2 \\ 
		DKRL  & - & - & 40.0 & 23.0$\;$ & - & - & - & - & - & - & - & -\\ %\hline
		ConMask & 47.1 & 64.5 & \textbf{81.0} & 58.4* & 21.5 & 39.9 & 45.8 & 29.9 & 42.3 & \textbf{57.3} & \textbf{71.7} & \textbf{53.3} \\ 
		%ConMask & 47.1 & 64.5 & 81.0 & 61.0 & 21.5 & 39.9 & 45.8 & 48.6 & ? & ? & 71.7 & 53.3 \\ \hline 
		Cmplx-OWE-200 & 49.0 & 62.3 & 73.6 & 57.7$\;\;$ & 29.1 & 41.0 & 52.7 & 37.3 & 44.2 & 55.9 & 68.2 & 52.3 \\
		Cmplx-OWE-300 & \textbf{51.9} & \textbf{65.2} & 76.0 & \textbf{60.3}$\;\;$ & \textbf{31.6} & \textbf{43.9} & \textbf{56.0} & \textbf{40.1} & \textbf{44.8} & 57.1 & 69.1 & 53.1 \\
		\bottomrule
	\end{tabular}
	\caption{Comparison with other open-world KGC models on tail prediction. Note that we used the same evaluation protocol with target filtering as in ConMask. The asterisk (*) denotes that the result differs from the one published, because the MRR is calculated differently.}
	\label{table:comparison_with_conmask}
\end{table*}

\subsection{Experimental Setup} \label{sec:experimental_setup}

%We train a closed-world link prediction model 
We perform multiple runs using different KGC models, transformation types, training data, and embeddings used. For each run, both KGC model and transformation $\Psi^{map}$ are trained on the training set: the KGC model without using any textual information and the transformation using entity names and descriptions. We manually optimize all hyperparameters on the validation set. Due to the lack of an open-world validation set on FB20k, we randomly sampled 10\% of the test triples as a validation set.

\subsubsection {Performance Measures}
Performance figures are computed using tail prediction on the test sets: For each test triple $(h,r,t)$ with open-world head $h \notin E$, we rank all known entities $t' \in E$ by their score $\phi(h,r,t')$. We then evaluate the ranks of the target entities $t$ with the commonly used mean rank (MR), mean reciprocal rank (MRR), as well as Hits@1, Hits@3, and Hits@10.

Note that multiple triples with the same head and relation but different tails may occur in the dataset: $(h,r,t_1),...,(h,r,t_p)$. Following~\cite{bordes2013transe}, when evaluating triple $(h,r,t_i)$ we remove all entities $t_1,...,t_{i-1},t_{i+1},...,t_p$ from the result list . All results (except {\it MRR(raw)}) are reported with this {\it filtered} approach.
Note also that when computing the MRR, given a triple $(h,r,t_i)$ only the reciprocal rank of $t_i$ itself is evaluated (and {\it not} the best out of $t_1,...,t_{i},...,t_p$, which would give better results). This is common when evaluating KGC models~\cite{bordes2013transe} but differs from ConMask's evaluation code, which is why one result in Table \ref{table:comparison_with_conmask} differs from~\cite{shi17openworld} (see the (*) mark).

Note also that~\cite{shi17openworld} add a second filtering method called {\it target filtering}: When evaluating a test triple $(h,r,t)$, tails $t'$ are only included in the ranked result list if a triple of the form $(?,r,t')$ exists in the training data, otherwise it is skipped. We found this to improve quantitative results substantially, but it limits the predictive power of the model because tails can never be linked via {\it new} relations. Therefore, we use target filtering only when comparing with the ConMask and DKRL models from~\cite{shi17openworld} (Table \ref{table:comparison_with_conmask}).
%we use the target filtering (different from the \textit{filtering} above) method described in Shi et al. \cite{shi17openworld}. Thereby we  compute scores only for those entities, that appear with the same relation at target position within the training set. Note that in this method test triples for which the target does not appear in the resulting list of entities are skipped. 
%In contrast to the evaluation method of ConMask we compute the MRR on the resulting list of ranks and do not group the result by relation 

% To mitigate the excessive cost involved in computing
% scores for all entities in the KG, we applied a target filtering
% method to all KGC models. Namely, for a given partial
% triple hh, r, ?i or h?, r, ti, if a target entity candidate has
% not been connected via relationship r before in the training
% set, then it is skipped, otherwise we use the KGC model to
% calculate the actual ranking score. Simply put, this removes
% relationship-entity combinations that have never before been
% seen and are likely to represent nonsensical statements

\subsubsection{Implementation Details}
For training TransE and DistMult, we use the OpenKE framework\footnote{OpenKE framework: https://github.com/thunlp/OpenKE} which provides implementations of many common link prediction models. For closed-world graph embedding, we use both OpenKE and our own implementation after validating the equivalence of both.

%We optimized the hyperparameters using the validation set. 
For training the transformation $\Psi^{map}$, we used the Adam optimizer with a learning rate of $10^{-3}$ and batch size of $128$. For DBPedia50k we use a dropout of 0.5, while for FB20k and FB15k-237-OWE we use no dropout. The embedding used is the pretrained 300 dimensional Wikipedia2Vec embedding and the transformation used is affine unless stated otherwise. 

%\subsection{Results}
\begin{table}[hb]
	\centering
	\begin{tabular}{@{\extracolsep{0.8ex}}l r r r r r} 
	\toprule \\[-0.8em]
	& \multicolumn{2}{c}{MRR} & \multicolumn{3}{c}{HITS@}  \\ 
	\cline{2-3} \cline{4-6}\\[-0.8em]
	\multicolumn{1}{c}{Model} & Filt. & Raw & 1 & 3 & 10 \\
		\midrule \\[-0.8em]
		TransE-OWE & 28.7 & 22.9 & 21.9 & 31.7 & 41.0 \\  
		DistMult-OWE & 34.4 & 25.7 & 26.6 & 37.7 & \textbf{49.2} \\
		ComplEx-OWE & \textbf{35.2} & \textbf{26.1} & \textbf{27.8} & \textbf{38.6} & 49.1 \\
		\bottomrule
	\end{tabular}
	\caption{Open-world tail prediction by applying the transformation to different closed-world link prediction models on the FB15k-237-OWE dataset without target filtering.}
	\label{table:different_models}
\end{table}

\begin{table}[ht]
\centering
\begin{tabular}{@{\extracolsep{0.75ex}}l r r r r r } 
	\toprule \\[-0.8em]
	& \multicolumn{2}{c}{MRR} & \multicolumn{3}{c}{HITS@}  \\ 
	\cline{2-3} \cline{4-6}\\[-0.8em]
	\multicolumn{1}{c}{Transformation} & Filt. & Raw & 1 & 3 & 10 \\
	\hline \\[-0.8em]
	Linear & 33.2 & 25.2 & 26.1 & 36.4 & 46.5 \\ 
	Affine & \textbf{35.2} & \textbf{26.1} & \textbf{27.8} & \textbf{38.6} & \textbf{49.1} \\
	MLP & 32.2 & 24.7 & 25.0 & 35.3 & 45.6 \\
	\bottomrule
\end{tabular}
\caption{Comparison of different transformation functions with ComplEx-OWE-300 on the FB15k-237-OWE dataset without target filtering.}
\label{table:transformations}
\end{table}
\begin{figure*}[ht!]
\centering
\begin{tabular}{cc}
    \includegraphics[width=0.45\textwidth]{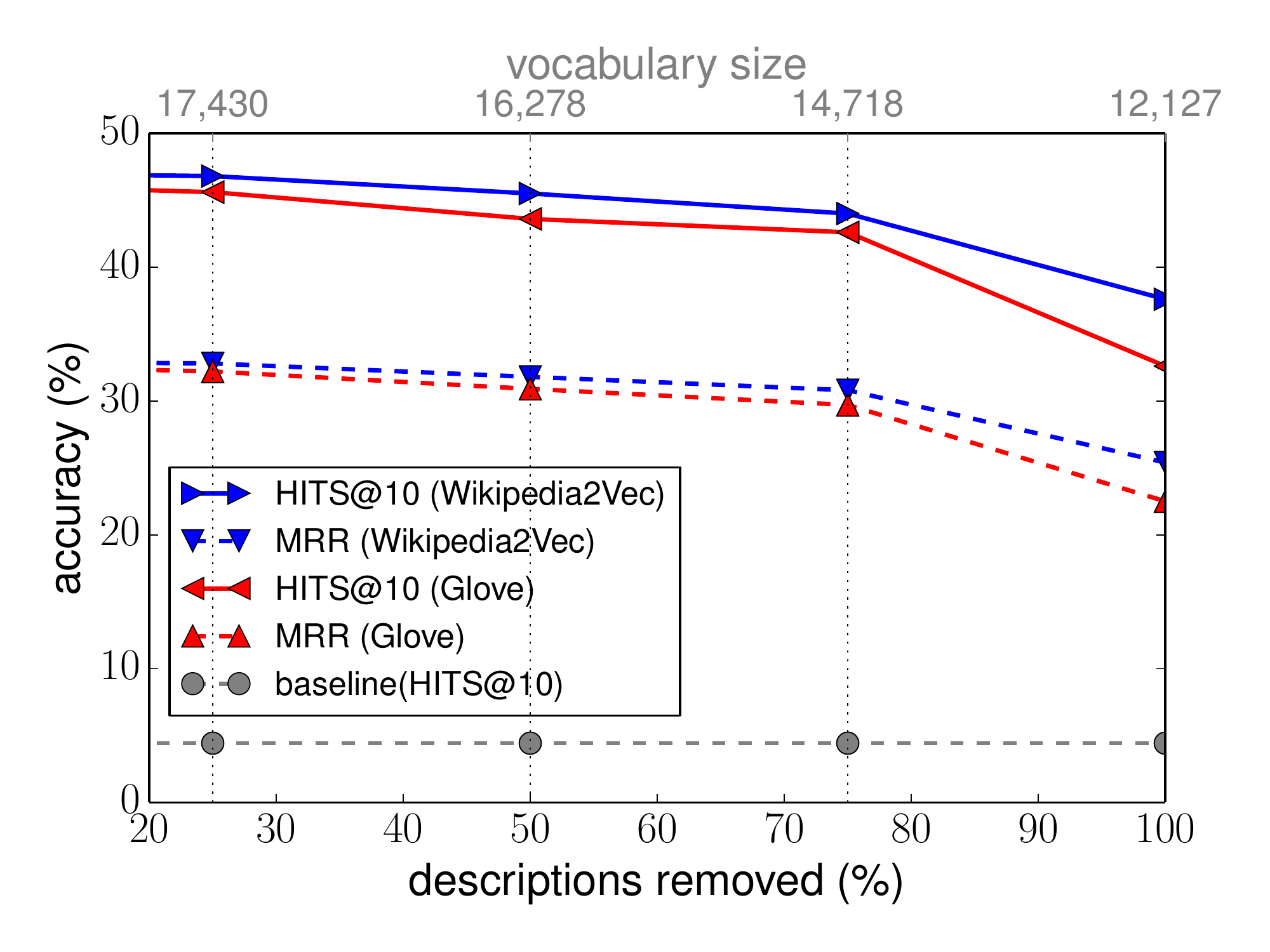} &
    \includegraphics[width=0.45\textwidth]{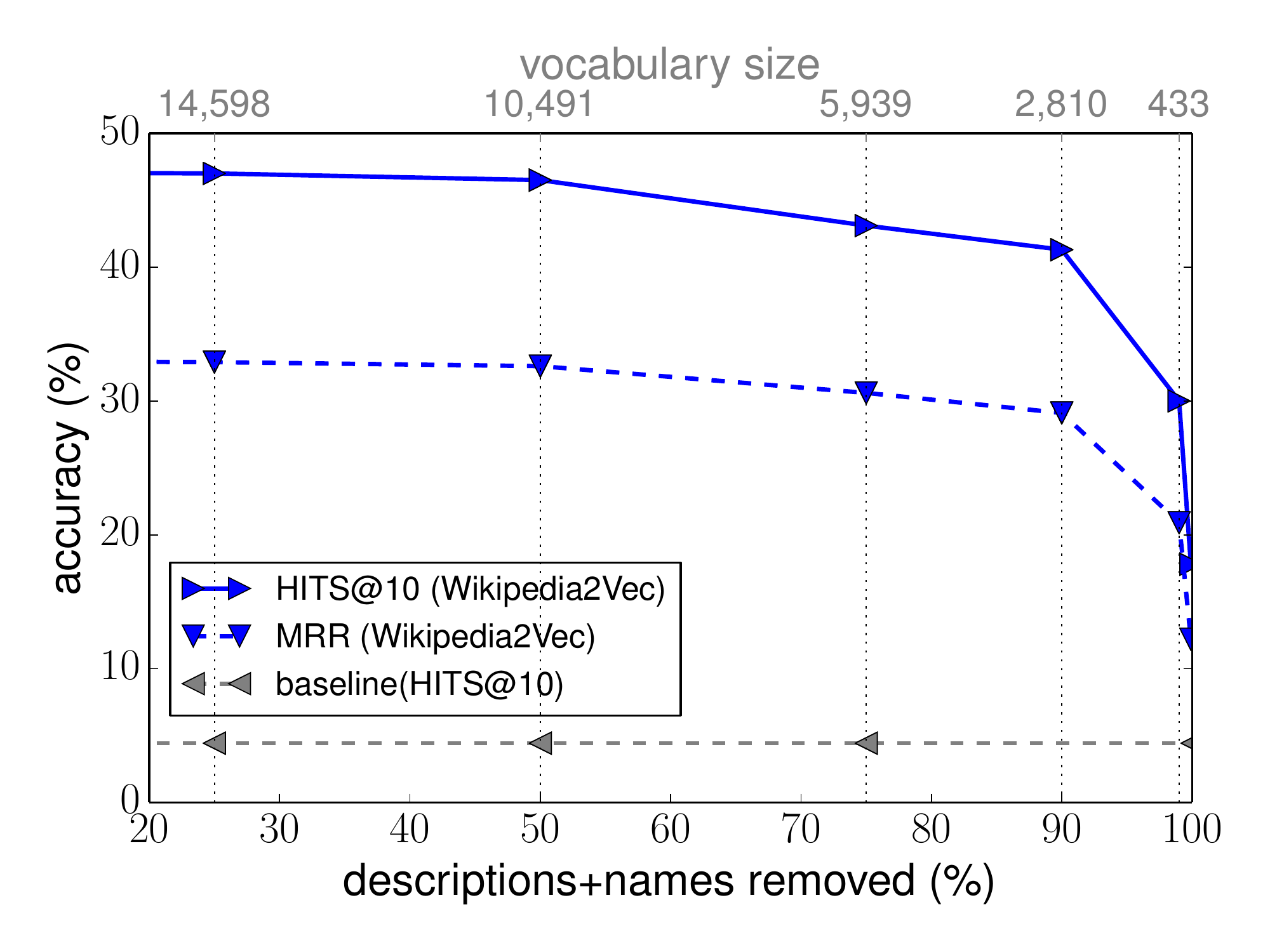}\\[-0.5em]
    (a) Scarce Entity Descriptions & (b) Scarce Entity Metadata (Names and Descriptions)
\end{tabular}\\[-0.5em]
\caption{Performance on FB15k-237-OWE with ComplEx-OWE-300 without target filtering when dropping (a) the entity descriptions or (b) both descriptions and names. The x-axis shows the amount of textual data removed. Even for scarce textual data, learning the transformation $\Psi^{map}$ is robust.}
\label{fig:scarce}
\end{figure*}

\subsection{Comparison with State of the Art}

We first compare our model ComplEx-OWE with other open-world link prediction models in Table \ref{table:comparison_with_conmask}. For a fair comparison, all the results are evaluated using target filtering. For all models and all datasets, 200-dimensional Glove embeddings were used,  except for the Complex-OWE300, which uses 300-dimensional Wikipedia2Vec embeddings. The effect of different embeddings will be studied further in Section \ref{sec:robustness}. 

The results for Target Filtering Baseline, DKRL and ConMask were obtained by the implementation provided by \cite{shi17openworld}. The Target Filtering Baseline is evaluated by assigning random scores to all targets that pass the target filtering criterion. DKRL uses a two-layer CNN over the entity descriptions. ConMask uses a CNN over the entity names and descriptions along with the relation-based attention weights.

It can be seen from Table \ref{table:comparison_with_conmask} that our best model, ComplEx-OWE300, performs competitively when compared to ConMask. On DBPedia50k, our model performs best on all metrics except Hits@10. On FB20k it is outperformed by a small margin by ConMask but performs better on Hits@1. 
%However, the difference of performance between ConMask and ComplEx-OWE300 on FB20k is very small. 
On FB15k-237-OWE our model outperforms all other models significantly. We believe that this is due to FB15k-237-OWE having very short descriptions. ConMask generally relies on extracting information from the description of entities with its attention mechanism, whereas our model relies more on extracting information from the textual corpus that the word embedding were trained on. This enables our model to provide good results without relying on having long descriptions.

%
%\subsubsection {Comparison of encoders }
%\begin{table*}[h]
%	\centering
%	\begin{tabular}{@{\extracolsep{1ex}}c r r r r r r r r r r r r r r r} 
%		\hline \\[-0.8em]
%		& \multicolumn{5}{c}{DBPedia50k} & \multicolumn{5}{c}{DBPedia500k} & \multicolumn{5}{c}{FB15k-237-OWE} \\
%		\cline{2-6} \cline{7-11} \cline{12-16} \\[-0.8em]
%		& \multicolumn{2}{c}{MRR} & \multicolumn{3}{c}{HITS@} & \multicolumn{2}{c}{MRR} & \multicolumn{3}{c}{HITS@} & \multicolumn{2}{c}{MRR} & \multicolumn{3}{c}{HITS@}  \\ 
%		\cline{2-3} \cline{4-6}  \cline{7-8} \cline{9-11} \cline{12-13}  \cline{14-16}  \\[-0.8em]
%		Model & Filtered & Raw & 1 & 3 & 10 & Filtered & Raw & 1 & 3 & 10 & Filtered & Raw & 1 & 3 & 10\\
%		\hline \\[-0.8em]
%		Average & ? & ? & ? & ? & ? & ? & ? & ? & ? & ? & ? & ? & ? & ? & ?\\ 
%		LSTM & ? & ? & ? & ? & ? & ? & ? & ? & ? & ? & ? & ? & ? & ? & ? \\
%		Bi-LSTM & ? & ? & ? & ? & ? & ? & ? & ? & ? & ? & ? & ? & ? & ? & ? \\
%		\hline
%	\end{tabular}
%	\caption{Encoder comparison}
%	\label{table:2}
%\end{table*}
%
%We show the effect of using different encoders

\subsection {Analysis of Different Link Prediction Models and Transformations}

Our OWE extension for open-world link prediction can be used with any common KGC model. 
Therefore, we evaluate three commonly used options, namely TransE, DistMult, and ComplEx. Results are displayed in Table \ref{table:different_models}: All three models are trained with embedding dimensionality $d=300$ on the closed-world dataset. For text embeddings, Wikipedia2Vec embeddings of the same dimensionality were used. It can be seen that the performance on the open-world setting matches the expressiveness of the models: ComplEx-OWE with its ability to model even asymmetric relations yields the best results, while the symmetric DistMult-OWE achieves a similar performance. 
%Overall, the performance of our model depends on the expressiveness of the underlying KGC model.

We also test different transformation functions $\Psi^{map}$ as illustrated in Table \ref{table:transformations}. It can be observed that quite simple transformations achieve the strong results: The best performance is achieved by the affine transformation with $49.1$\% HITS@10 by a margin of 2--4 percent. 

\begin{table*}[t]
    \centering
    \begin{tabular}{p{4cm}p{4cm}p{4cm}p{4cm}}
    \toprule
    Test Triple & Head Description & Top 4 Nearest Neighbors & Top 4 Predictions \\
    \midrule
        %%%%%%%%%%%%%%%%%%%%%%%%%%%%%
    ({\bf Bram Stoker},\newline
     /people/person/profession,\newline
     \textbf{Writer}) & 
     Irish novelist and short story writer, best known today for his 1897 Gothic novel Dracula
     &
     1. Ursula K. Le Guin \newline
     2. Charles Stross \newline
     3. Larry Niven \newline     
     4. Kurt Vonnegut 
     %5. major league baseball allstar game
         &  
    {\bf 1. Writer}\newline
    2. England\newline
    3. United Kingdom\newline
    4. Author
    %5. italy         
        \\
         \midrule         
    % 
    %%%%%%%%%%%%%%%%%%%%%%%%%%%%%
    ({\bf Parma},\newline
     /.../country,\newline
     \textbf{Italy}) & 
     Italian comune
     &
     1. Essen \newline
     2. Mantua \newline
     3. Bergamo \newline     
     4. Siena 
     %5. keble college
         &  
    1. Portugal\newline
    2. Netherlands\newline
    3. Germany\newline
    {\bf 4. Italy}
    %5. italy         
         \\ \midrule
    %
    %%%%%%%%%%%%%%%%%%%%%%%%%%%%%
    ({\bf Bachelor of Science},\newline
     /.../institution,\newline
     \bf Kingston University) & 
     Academic degree &
     1. Masters Degree \newline
     2. Doctoral Degree \newline
     3. Bachelor of Arts \newline     
     4. Master of Arts 
     %5. bachelor of business administration
         &  
    1. Harvard Law School\newline
    2. Wesleyan University\newline
    3. Panjab University\newline
    4. Baylor University\vspace{0.1cm}\newline
    \textbf{32. Kingston University}\\ \midrule  
    %5. wellesley college
        % 
    %%%%%%%%%%%%%%%%%%%%%%%%%%%%%
    ({\bf Amtrak},\newline
     /.../headquarters.../citytown,\newline
     \bf Washington DC)  & 
     Intercity rail operator in the United States &
     1. AT\&T \newline
     2. Southwest Airlines \newline
     3. Starbucks \newline     
     4. Delta Airlines
     %5. associated press
         &  
    1. Dublin\newline
    2. New York City\newline
    3. United States Dollar\newline
    4. Charlotte\vspace{0.1cm}\newline
    \textbf{72. Washington DC}
    %5. forward         
         \\ 
             \bottomrule

    \end{tabular}
    \caption{Selected results on FB15k-237-OWE. For each test triple (Column 1), the head's name and description (Column 2) is mapped to graph-based embedding space. The nearest training entities in that space (Column 3) indicate a good semantic match. The model predicts reasonable tails, in the first two cases successfully, in the others not. }
    \label{tab:sampleresults}
\end{table*}

\subsection{Text Embeddings and Robustness To Missing Entity Metadata}
\label{sec:robustness}

In some cases, the knowledge graph may lack textual metadata (both the name and description) for some or all of its entities. Other models like ConMask and DKRL are dependant on textual descriptions, e.g. ConMask uses attention mechanisms to select relation-specific target words from long texts.
Therefore, ConMask and DKRL would require completely dropping triples without metadata and be unable to learn about the link structure of such entities as they use joint training. However, in our approach, we have to drop such entities only during the phase where the transformation $\Psi^{map}$ is learned, while the link prediction model can still be learned on the full graph. 

To demonstrate the robustness of our approach to missing entity meta-data, we re-evaluate accuracy when randomly dropping metadata for training entities. Fig. \ref{fig:scarce} outlines the performance for two scenarios:
\begin{itemize}
    \item {\bf Dropping descriptions}: We remove only the textual descriptions for a varying percentage of randomly selected entities (between $20$\% to $100$\%). The names of these entities are not removed and therefore, we still train $\Psi^{map}$ on them.
    \item {\bf Dropping all meta-data}: We randomly select entities and remove both their descriptions and names, effectively removing these entities from the training set altogether when training $\Psi^{map}$.
\end{itemize}
We also included a baseline experiment to simulate an unsuccessful learning of $\Psi^{map}$. In this baseline, when evaluating a test triple, we replace its head by the embedding of another random head from the training data. Note that this baseline still gives some reasonable hits for triples where the {\it relation} is a strong indicator. For example, if we have a triplet $(X,time\_zone,?)$: Even if the head $X$ is unknown, a model can achieve reasonable accuracy by simply ''guessing'' time zones as tails.

Overall, Fig. \ref{fig:scarce} suggests that transformation learning is able to generalize well even with very limited training data.
In Fig. \ref{fig:scarce}a only the descriptions of entities have been removed. For Wikipedia2Vec embeddings, this removal has virtually no effect on prediction accuracy. We believe that this is because Wikipedia2Vec embeddings are trained such that we can lookup strong entity embeddings by the name alone. Even when removing 100\% of descriptions (i.e., only training on the entity names), accuracy is only 2-3\% lower than training on the full graph. However, in case of Glove embeddings, the drop in performance is very significant, especially when the description is dropped for all the entities.

In Fig. \ref{fig:scarce}b, we remove not only descriptions but also entity names. Even in this case, learning is robust. If half of the $12,324$ training entities are removed, the drop in accuracy is less than $1$\%. Only when removing $90$\% of training data (leaving $123$ training entities), performance starts to deteriorate significantly.
This highlights the ability of our model to learn from a limited amount of training data, when it is important to be able to train the KGC model itself on {\it all} the entities.

%%%%%%%%%%%%%%%%%%%%%%%%%%%%%%%%%%%%%%%%%%%%

\subsection{Selected Results}
Finally, we inspect sample prediction results for ComplEx-OWE-300 in Table \ref{tab:sampleresults}. Besides the final prediction, we also test whether our transformation from text-based to semantic space is successful: For each test triple, we represent the open-world head entity by its text-based embedding $v_{head}$, match it to a graph-based embedding $\Psi^{map}(v_{head})$, and estimate the nearest neighbor entities in this space. We use the Euclidean distance on the real part of the ComplEx embeddings, but found results to be similar for the imaginary part.

If the transformation works well, we expect these nearest neighbors to be semantically similar to the head entity. This is obviously the case: For {\it Bram Stoker} (the author of Dracula), the nearest neighbors are other authors of fantasy literature. For {\it Parma}, the neighbors are cities (predominantly in Italy). For {\it Bachelor of Science}, the model predicts appropriate entities (namely, Universities) but -- even though we apply filtering -- the predictions are not rewarded. This is because the corresponding triples, like {\it (Bachelor of Science, /.../institution, Harward Law School)}, are missing in the knowledge graph. 

%Given this reasonable entity embedding, the graph-based model makes reasonable predictions in most cases. 
%Find nearest neighbors to $\Psi^{map}(v_{head}$
%using Euclidean distance.

%Complex model, real-valued embeddings (similar results with imaginary parts).

%%%%%%%%%%%%%%%%%%%%%%%%%%%%%%%%%%%%%%%%%%%%%

\section{Conclusion}
In this work, we have presented a simple yet effective extension to embedding-based knowledge graph completion models (such as ComplEx, DistMult and TransE) to perform open-world prediction. Our approach -- which we named OWE -- maps text-based entity descriptions (learned from word embeddings) to the pre-trained graph embedding space. In experiments on several datasets (including the new FB15K-237-OWE dataset we introduced in this work), we showed that the learned transformations yield semantically meaningful results, that the approach performs competitive with respect to the state of the art, and that it is robust to scarce text descriptions.

An interesting direction of future work will be to combine our model with approaches like ConMask~\cite{shi17openworld}, which (1) exploit more complex aggregation functions and (2) use relation-specific attention/content masking to draw more precise embeddings from longer descriptions.

\section*{Acknowledgements}
This work was partially funded by the German Federal Ministry of Education and Research (Program FHprofUnt, Project DeepCA / 13FH011PX6) and the German Academic Exchange Service (Project FIBEVID  / 57402798).

%References and End of Paper
%These lines must be placed at the end of your paper
\bibliography{refs}

\begin{thebibliography}{}

\bibitem[\protect\citeauthoryear{Bojanowski \bgroup et al\mbox.\egroup
  }{2017}]{bojanowski16enriching}
Bojanowski, P.; Grave, E.; Joulin, A.; and Mikolov, T.
\newblock 2017.
\newblock {Enriching Word Vectors with Subword Information}.
\newblock {\em {TACL}} 5:135--146.

\bibitem[\protect\citeauthoryear{Bordes \bgroup et al\mbox.\egroup
  }{2013}]{bordes2013transe}
Bordes, A.; Usunier, N.; Garcia-Duran, A.; Weston, J.; and Yakhnenko, O.
\newblock 2013.
\newblock {Translating Embeddings for Modeling Multi-relational Data}.
\newblock In {\em {Adv. in Neural Information Processing Systems}},
  2787--2795.

\bibitem[\protect\citeauthoryear{Dong \bgroup et al\mbox.\egroup
  }{2014}]{dong14knowledgevault}
Dong, X.~L.; Gabrilovich, E.; Heitz, G.; Horn, W.; Lao, N.; Murphy, K.;
  Strohmann, T.; Sun, S.; and Zhang, W.
\newblock 2014.
\newblock {Knowledge Vault: A Web-Scale Approach to Probabilistic Knowledge
  Fusion}.
\newblock In {\em Proc. KDD},  601--610.

\bibitem[\protect\citeauthoryear{Faruqui \bgroup et al\mbox.\egroup
  }{2015}]{faruqui14retrofitting}
Faruqui, M.; Dodge, J.; Jauhar, S.~K.; Dyer, C.; Hovy, E.~H.; and Smith, N.~A.
\newblock 2015.
\newblock {Retrofitting Word Vectors to Semantic Lexicons}.
\newblock In {\em NAACL HLT},  1606--1615.

\bibitem[\protect\citeauthoryear{Ferrucci \bgroup et al\mbox.\egroup
  }{2010}]{ferrucci10watson}
Ferrucci, D.; Brown, E.; Chu-Carroll, J.; Fan, J.; Gondek, D.; Kalyanpur,
  A.~A.; Lally, A.; Murdock, J.~W.; Nyberg, E.; Prager, J.; Schlaefer, N.; and
  Welty, C.
\newblock 2010.
\newblock {Building Watson: An Overview of the DeepQA Project}.
\newblock {\em AI Magazine} 31(3):59--79.

\bibitem[\protect\citeauthoryear{Goldberg}{2016}]{goldberg16primer}
Goldberg, Y.
\newblock 2016.
\newblock {A Primer on Neural Network Models for Natural Language Processing}.
\newblock {\em J. Artif. Int. Res.} 57(1):345--420.

\bibitem[\protect\citeauthoryear{Mikolov \bgroup et al\mbox.\egroup
  }{2013}]{mikolov13w2v}
Mikolov, T.; Sutskever, I.; Chen, K.; Corrado, G.; and Dean, J.
\newblock 2013.
\newblock Distributed representations of words and phrases and their
  compositionality.
\newblock {\em CoRR} abs/1310.4546.

\bibitem[\protect\citeauthoryear{Nickel \bgroup et al\mbox.\egroup
  }{2016}]{nickel16kg}
Nickel, M.; Murphy, K.; Tresp, V.; and Gabrilovich, E.
\newblock 2016.
\newblock {A Review of Relational Machine Learning for Knowledge Graphs}.
\newblock {\em Proceedings of the IEEE} 104(1):11--33.

\bibitem[\protect\citeauthoryear{Paulheim}{2017}]{paulheim17kgrefinement}
Paulheim, H.
\newblock 2017.
\newblock {Knowledge Graph Refinement: {A} Survey of Approaches and Evaluation
  Methods}.
\newblock {\em Semantic Web} 8(3):489--508.

\bibitem[\protect\citeauthoryear{Pennington, Socher, and
  Manning}{2014}]{Pennington14glove:global}
Pennington, J.; Socher, R.; and Manning, C.~D.
\newblock 2014.
\newblock {Glove: Global Vectors for Word Representation}.
\newblock In {\em Proc. EMNLP}.

\bibitem[\protect\citeauthoryear{Peters \bgroup et al\mbox.\egroup
  }{2018}]{peters18elmo}
Peters, M.~E.; Neumann, M.; Iyyer, M.; Gardner, M.; Clark, C.; Lee, K.; and
  Zettlemoyer, L.
\newblock 2018.
\newblock {Deep contextualized word representations}.
\newblock In {\em Proc. NAACL}.

\bibitem[\protect\citeauthoryear{Ristoski and
  Paulheim}{2016}]{ristoski16rdf2vec}
Ristoski, P., and Paulheim, H.
\newblock 2016.
\newblock {RDF2Vec: {RDF} Graph Embeddings for Data Mining}.
\newblock In {\em Proc. ISWC},  498--514.

\bibitem[\protect\citeauthoryear{Shi and Weninger}{2017a}]{shi17openworld}
Shi, B., and Weninger, T.
\newblock 2017a.
\newblock {Open-World Knowledge Graph Completion}.
\newblock {\em CoRR} abs/1711.03438.

\bibitem[\protect\citeauthoryear{Shi and Weninger}{2017b}]{shi16proje}
Shi, B., and Weninger, T.
\newblock 2017b.
\newblock {ProjE: Embedding Projection for Knowledge Graph Completion}.
\newblock In {\em Proc. {AAAI}},  1236--1242.

\bibitem[\protect\citeauthoryear{Singhal}{2012}]{gogle12knowledgegraph}
Singhal, A.
\newblock 2012.
\newblock {Introducing the Knowledge Graph: Things, not Strings}.
\newblock https://googleblog.blogspot.com
  /2012/05/introducing-knowledge-graph-things-not.html, last retrieved: Aug
  2018).

\bibitem[\protect\citeauthoryear{Socher \bgroup et al\mbox.\egroup
  }{2013}]{socher13ntn}
Socher, R.; Chen, D.; Manning, C.~D.; and Ng, A.~Y.
\newblock 2013.
\newblock Reasoning with neural tensor networks for knowledge base completion.
\newblock In {\em Proceedings of the 26th International Conference on Neural
  Information Processing Systems - Volume 1}, NIPS'13,  926--934.
\newblock USA: Curran Associates Inc.

\bibitem[\protect\citeauthoryear{Toutanova and
  Chen}{2015}]{toutanova2015fb15k237}
Toutanova, K., and Chen, D.
\newblock 2015.
\newblock {Observed Versus Latent Features for Knowledge Base and Text
  Inference}.
\newblock In {\em 3rd Workshop on Continuous Vector Space Models and Their
  Compositionality}.

\bibitem[\protect\citeauthoryear{Trouillon \bgroup et al\mbox.\egroup
  }{2016}]{trouillon2016complex}
Trouillon, T.; Welbl, J.; Riedel, S.; Gaussier, {\'E}.; and Bouchard, G.
\newblock 2016.
\newblock {Complex Embeddings for Simple Link Prediction}.
\newblock In {\em Int. Conference on Machine Learning},  2071--2080.

\bibitem[\protect\citeauthoryear{Wang and Li}{2016}]{wang16teke}
Wang, Z., and Li, J.
\newblock 2016.
\newblock {Text-enhanced Representation Learning for Knowledge Graph}.
\newblock In {\em Proc. International Joint Conference on Artificial
  Intelligence},  1293--1299.

\bibitem[\protect\citeauthoryear{Xie \bgroup et al\mbox.\egroup
  }{2016}]{xie16descriptions}
Xie, R.; Liu, Z.; Jia, J.; Luan, H.; and Sun, M.
\newblock 2016.
\newblock {Representation Learning of Knowledge Graphs with Entity
  Descriptions}.
\newblock In {\em Proc. AAAI},  2659--2665.

\bibitem[\protect\citeauthoryear{Xu \bgroup et al\mbox.\egroup
  }{2014}]{xu14rcnet}
Xu, C.; Bai, Y.; Bian, J.; Gao, B.; Wang, G.; Liu, X.; and Liu, T.-Y.
\newblock 2014.
\newblock {RC-NET: A General Framework for Incorporating Knowledge into Word
  Representations}.
\newblock In {\em Proc. Int. Conf. on Information and Knowledge Management},
  1219--1228.

\bibitem[\protect\citeauthoryear{Xu \bgroup et al\mbox.\egroup
  }{2017}]{xu17jointly}
Xu, J.; Qiu, X.; Chen, K.; and Huang, X.
\newblock 2017.
\newblock {Knowledge Graph Representation with Jointly Structural and Textual
  Encoding}.
\newblock In {\em Proc. Int. Joint Conference on Artificial Intelligence},
  1318--1324.

\bibitem[\protect\citeauthoryear{Yamada \bgroup et al\mbox.\egroup
  }{2016}]{wikipedia2vec}
Yamada, I.; Shindo, H.; Takeda, H.; and Takefuji, Y.
\newblock 2016.
\newblock {Joint Learning of the Embedding of Words and Entities for Named
  Entity Disambiguation}.
\newblock In {\em Proc. SIGNLL Conference on Computational Natural Language
  Learning},  250--259.

\bibitem[\protect\citeauthoryear{Yang \bgroup et al\mbox.\egroup
  }{2014}]{yang2015distmult}
Yang, B.; Yih, W.; He, X.; Gao, J.; and Deng, L.
\newblock 2014.
\newblock {Embedding Entities and Relations for Learning and Inference in
  Knowledge Bases}.
\newblock {\em CoRR} abs/1412.6575.

\bibitem[\protect\citeauthoryear{Yu and Dredze}{2014}]{yu14improving}
Yu, M., and Dredze, M.
\newblock 2014.
\newblock {Improving Lexical Embeddings with Semantic Knowledge}.
\newblock In {\em Proc. ACL},  545--550.

\end{thebibliography}
\bibliographystyle{aaai}

\end{document}